**Inverse design optimization framework via a two-step deep learning approach: application to a wind turbine airfoil**

Sunwoong Yang[a], Sanga Lee[b], Kwanjung Yee[a*]

[a]Department of Aerospace Engineering, Seoul National University, Seoul 08826, Republic of Korea

[b]Korea Institute of Industrial Technology, Incheon 21999, Republic of Korea

***Corresponding author**

Email: kjyee@snu.ac.kr

**ORCID**

Sunwoong Yang: 0000-0003-2065-9139

Sanga Lee: 0000-0002-6567-0502

Kwanjung Yee: 0000-0001-9384-7566

**Acknowledgments**

This work was supported by the National Research Foundation of Korea (NRF) Grant funded by the Ministry of Science and ICT (NRF-2017R1A5A1015311).

**Abstract**

The inverse approach is computationally efficient in aerodynamic design as the desired target performance distribution is specified. However, it has some significant limitations that prevent it from achieving full efficiency. First, the iterative procedure should be repeated whenever the specified target distribution changes. Target distribution optimization can be performed to clarify the ambiguity in specifying this distribution, but several additional problems arise in this process such as loss of the representation capacity due to parameterization of the distribution, excessive constraints for a realistic distribution, inaccuracy of quantities of interest due to theoretical/empirical predictions, and the impossibility of explicitly imposing geometric constraints. To deal with these issues, a novel inverse design optimization framework with a two-step deep learning approach is proposed. A variational autoencoder and multi-layer perceptron are used to generate a realistic target distribution and predict the quantities of interest and shape parameters from the generated distribution, respectively. Then, target distribution optimization is performed as the inverse design optimization. The proposed framework applies active learning and transfer learning techniques to improve accuracy and efficiency. Finally, the framework is validated through aerodynamic shape optimizations of the wind turbine airfoil. Their results show that this framework is accurate, efficient, and flexible to be applied to other inverse design engineering applications.

**Keywords:** inverse design optimization, multi-layer perceptron, variational autoencoder, wind turbine airfoil

**Article highlights**

- This article proposed an inverse design optimization framework with a two-step deep learning approach to overcome the limitations of the previous inverse design studies

- The proposed framework was validated through two aerodynamic shape optimization problems of the wind turbine airfoil: single-objective and multi-objective optimizations

- Finally, their results confirmed the accuracy, efficiency, and flexibility of the proposed inverse design optimization framework via a two-step deep learning approach

# 1 Introduction

Recent advances in high-performance computing have enabled aerodynamic engineers to use high-fidelity analyses, offering a wide range of options in the aerodynamic design process. Accordingly, numerous novel design methodologies have been developed, most of which are based on two conventional aerodynamic design methods: inverse and direct design approaches [1-3]. In particular, inverse design is computationally efficient in that the desired target performance distribution is already defined and the corresponding design shape can be calculated with a few iterations coupled with a flow solver [4-6].

However, the inverse method also has a critical disadvantage: whenever the target distribution changes, an iterative process to find the design shape matching the target distribution should be repeated. Considering that most design stages require significant trial and error, this process undermines the efficiency of the inverse design. Therefore, several researchers have used a surrogate model in inverse design to avoid this iterative process. In particular, artificial neural network (ANN) surrogate models have been widely used owing to their universal approximation property [7]. Kharal and Saleem [8] and Sun et al. [9] used aerodynamic quantities of interest (QoI) as the inputs of an ANN model to obtain airfoil shape parameters as the output in an inverse design procedure. Wang et al. [10] also applied an ANN for the same purpose, but additionally performed dimensional reduction of input data to reduce the database size required for model training.

Though these studies do not require iterative procedures coupled with the flow solver, they still require the predefined performance distribution. For an efficient inverse design, an appropriate aerodynamic performance should be defined, which is highly dependent on the designer's engineering knowledge and experience. This ambiguity in specifying the target distribution has inspired researchers to optimize it. Obayashi and Takanashi [11] and Kim and Rho [12] used control points-based techniques to parameterize the pressure distributions, and then optimized the distribution using a genetic algorithm (GA). In these inverse design optimization processes, the aerodynamic QoI of the distribution were obtained through theoretical/empirical predictions, and a number of constraints were imposed to ensure the reality of the distribution. For instance, Obayashi and Takanashi [11] estimated the viscous drag using the Squire–Young empirical formula and imposed six constraints for realistic pressure coefficient ($C_p$) distributions. Though these studies attempted to deal with the fundamental limitation of the inverse design method, the following problems still exist: 1) loss of the diverse representation capacity of the $C_p$ distribution due to parameterization; 2) excessive constraints to ensure a realistic $C_p$ distribution; 3) discrepancies between the QoI predicted theoretically/empirically and those calculated using a flow solver; and 4) impossibility of explicitly imposing geometric constraints on the design shape.

To address the limitations of the representation capacity, QoI discrepancies, and excessive constraints, Zhu et al. [13] reduced the dimension of the $C_p$ distribution data via proper orthogonal decomposition (POD) and used the support vector regression (SVR) model to predict the aerodynamic performance of the airfoil from the reduced $C_p$ data. Then, a GA was implemented to optimize the $C_p$ distribution coupled with POD and SVR. Finally, the airfoil shape corresponding to the optimized pressure distribution was obtained in an iterative manner coupled with the flow solver. However, the limitation of geometric constraints has not been addressed since the prediction of the design shape is separated from the pressure optimization process. Therefore, when the shape predicted in the inverse design process violates geometric constraints, the design process should be traced back to the optimization process. Additionally, at the optimum solution, the discrepancy between the pressure distributions from prediction and calculation is noticeable, which indicates the low accuracy of its framework.

The drawbacks of the previous inverse design studies presented so far should be overcome by applying novel techniques in that computational efficiency, an essential advantage of inverse design, cannot be fully exploited. Therefore, this study proposes an inverse design optimization framework with a two-step deep learning approach. This approach refers to the sequential coupling of two deep learning models: variational autoencoder (VAE) [14] and multi-layer perceptron (MLP) [15]. The VAE and MLP were used to generate a realistic target distribution and to predict the QoI and shape parameters from the generated distribution, respectively. Then, target distribution optimization was performed as the inverse design optimization based on this approach. Active learning and transfer learning strategies were applied to improve the accuracy of the two-step approach-based optimization with efficient computational cost. The proposed inverse design optimization framework via a two-step deep learning approach was validated through aerodynamic shape optimization problems of the airfoil of a wind turbine blade, where inverse design is actively being applied.

This paper is organized as follows. *Section 2* describes the mathematical background of two deep learning models used in the inverse design optimization framework. *Section 3* presents the scheme of the proposed framework. In *section 4*, validation of the framework with application to a wind turbine airfoil is performed and the results are discussed. Finally, *section 5* concludes the study, emphasizing the flexibility of the presented framework.

## 2. Methodology

### 2.1 Multi-layer perceptron (MLP)

ANNs have been widely used in recent engineering design studies in that they can successfully learn complex nonlinear correlations between inputs and outputs [16, 17]. Furthermore, they are known for their efficiency in training high-dimensional problems with massive amounts of data. Among several ANN models, the MLP is the most common structure as it can learn various nonlinear physical phenomena using hidden layers; according to the universal approximation theorem, a neural network with a single hidden layer and finite nodes can approximate any continuous function [7]. In the inverse design optimization framework investigated in this study, an MLP model with these advantages was used as a regression surrogate model. This section presents a brief background of this model.

The MLP model passes on information received from the input layer to the output layer through the feed-forward process. Information transfer between each layer is performed through an affine transformation using weights and biases, as given in Eq. (1) (where $x$ is a vector of nodes at the current layer, $W$ and $b$ are the weight matrix and bias vector between the current layer and the next layer, respectively and $y$ is a vector of the nodes at the next layer). As it is based on a linear relationship, nonlinearity cannot be modeled no matter how many hidden layers are used.

$$y = Wx + b. \tag{1}$$

Modeling of nonlinearity is enabled by imposing activation functions on each layer. Typical examples of the functions are sigmoid, tanh, ReLU, and Leaky ReLU. The ReLU function has been used in numerous studies [18-20] because it has less saturation problems than sigmoid and tanh, and it does not perform expensive exponential operations. However, the ReLU function also has the disadvantage of being non-zero centered; therefore, the Leaky ReLU function is implemented as follows [21] ($a$ is for allowing a small, non-zero gradient and its value is set as 0.01 in this study):

$$f(x) = \begin{cases} x & if \; x > 0 \\ ax & otherwise \end{cases}.$$  (2)

The MLP model takes an activation function on $y$ in Eq. (1) and uses it as an input for the next layer to learn nonlinear behavior. This process is repeated until the data of the input layer reach the output layer. However, the final output will be quite different from the value of the training data as there is no training procedure for weights and biases in feed-forward. Accordingly, the weights and biases are trained based on the gradient of the loss function using backpropagation. They are updated through this process reducing the loss function of the model. Feed-forward (which transfers information from input to output) and backpropagation (which updates the model parameters) are repeated iteratively using gradient descent optimization techniques. Several optimization methods can be used, such as Adagrad [22], RMSprop [23], and Adam [24]. Adam has been widely used as it combines the strengths of Adagrad and RMSprop; its advantages are the ability to deal with sparse gradients and non-stationary objectives [24]. When the loss function decreases to the desired level, the optimization process ends and the final MLP model with fixed weights and biases serves as a surrogate model (prediction can be performed almost in real-time). Many papers adopting the MLP model have been published in various engineering fields; therefore, only the fundamental contents are presented in this paper. For more details on MLP, refer to Ref. [15].

### 2.2 Variational autoencoder (VAE)

A generative model learns how it is generated from real-world data [25]. In particular, a deep generative model uses neural networks with hidden layers to learn the underlying features of arbitrary data. Generative adversarial network (GAN) [26] and VAE [14] are typical examples of widely used deep generative models. Both models can be regarded as a dimensional reduction technique that learns a compact low-dimensional latent space from high-dimensional original data. Thus, the trained model can be used as a data generator in that it generates high-dimensional output data when it receives a low-dimensional latent variable. However, GAN has the disadvantage that it is difficult to generate continuous physical data such as pressure distribution and airfoil shape; therefore, auxiliary filters or layers should be used. For example, Chen et al. [27, 28] added a Bezier layer or a free-form-deformation layer at the end of the generator to enforce the continuity of the airfoil generated by GAN. In addition, Achour et al. [29] used the Savitzky–Golay filter on the generated airfoil for the same reason. However, when user-defined layers or filters are added to the neural network structure, additional hyperparameters are generated, thereby increasing the engineer's effort to train the model. Furthermore, GAN is known to be quite difficult to train owing to non-convergence, mode collapse, and vanishing gradients [30]. Therefore, in this study, a VAE was used as the data generator for the inverse design optimization. It is easy to train and does not require an extra post-processing technique to ensure continuity of the generated data. It will be shown later that the VAE used for this purpose was trained to generate smooth target distribution data for the inverse design.

The mathematical details of the VAE model are presented below. Note that this section is mainly based on Ref. [14].

Consider a dataset $X = \{x_i\}_{i=1}^N$, which consists of independent and identically distributed $N$ samples of variable $x$, and the unknown random variable $z$ is for generating $x$. When generating $x$ from $z$, variational inference is applied as the posterior $p_\theta(z|x)$ is intractable due to the intractability of the likelihood $p_\theta(z)$, where $\theta$ is a generative model parameter. Here, the approximation function $q_\phi(z|x)$ is used for the sampling instead of the intractable posterior $p_\theta(z|x)$, where $\phi$ is a variational parameter. Using this approximation, the marginal likelihood $p_\theta(x)$ can be expressed as

$$\log(p_\theta(x)) = \int \log(p_\theta(x)) \, q_\phi(z|x) dz \; \left( \because \int q_\phi(z|x) dz = 1 \right)$$

$$= \int \log\left( \frac{p_\theta(x,z)}{p_\theta(z|x)} \right) q_\phi(z|x) dz \; \left( \because p_\theta(x) = \frac{p_\theta(x,z)}{p_\theta(z|x)} \right)$$

$$= \int \log\left( \frac{p_\theta(x,z)}{q_\phi(z|x)} \cdot \frac{q_\phi(z|x)}{p_\theta(z|x)} \right) q_\phi(z|x) dz$$

$$= \int \log\left( \frac{p_\theta(x,z)}{q_\phi(z|x)} \right) q_\phi(z|x) dz + \int \log\left( \frac{q_\phi(z|x)}{p_\theta(z|x)} \right) q_\phi(z|x) dz. \tag{3}$$

The second term in the last equation is the Kullback–Leibler (KL) divergence of $q_\phi(z|x)$ from $p_\theta(z|x)$, $KL(q_\phi(z|x)||p_\theta(z|x))$, and it is impossible to calculate this term. Furthermore, because this term is always non-negative by its definition, the first term in the last equation becomes the lower bound on the marginal likelihood. Therefore, the maximization of the marginal likelihood can be replaced by the problem of maximizing the lower bound according to the evidence lower bound. The corresponding lower bound term can be divided as follows:

$$\int \log\left(\frac{p_\theta(x,z)}{q_\phi(z|x)}\right) q_\phi(z|x) dz$$

$$= \int \log\left(\frac{p_\theta(x|z)p_\theta(z)}{q_\phi(z|x)}\right) q_\phi(z|x) dz \;\; (\because p_\theta(x,z) = p_\theta(x|z)p_\theta(z))$$

$$= \int \log(p_\theta(x|z)) q_\phi(z|x) dz - \int \log\left(\frac{q_\phi(z|x)}{p_\theta(z)}\right) q_\phi(z|x) dz$$

$$= \mathbb{E}_{q_\phi(z|x)}[\log(p_\theta(x|z))] - \int \log\left(\frac{q_\phi(z|x)}{p_\theta(z)}\right) q_\phi(z|x) dz. \tag{4}$$

Therefore, maximization of the marginal likelihood $p_\theta(x)$ means maximizing the last equation of Eq. (4), where the first term is the reconstruction error of x through the posterior distribution (approximation function) $q_\phi(z|x)$ and the likelihood $p_\theta(x|z)$. The second term is the KL divergence of $q_\phi(z|x)$ from the prior $p_\theta(z)$, $KL(q_\phi(z|x)||p_\theta(z))$. In summary, the VAE model can be trained using the loss function consisting of the reconstruction error of data $x$ and the KL divergence of the approximation function from the prior.

In the derived loss function above, the approximation function $q_\phi(z|x)$ represents the dimensionality reduction procedure using an encoder (from original data to latent variable) and the likelihood $p_\theta(x|z)$ represents the dimensionality reconstruction procedure using a decoder (from latent variable to original data). To represent the intractability of these procedures, neural networks with hidden layers (MLP) are used to realize the encoder and decoder of the VAE. This explains why a VAE has a similar structure to a standard autoencoder (AE) [31]. However, the main difference is the latent space variable $z$. In an AE, the output of the encoder is used as a latent variable deterministically. On the other hand, the encoder of the VAE outputs a set of distribution parameters consisting of the mean ($\mu_i$) and standard deviation ($\sigma_i$) of the Gaussian distribution, and the latent space is sampled with randomness; to sample a $J$-dimensional latent variable $z$, the same number of sets $\{\mu_i, \sigma_i\}$ is required. However, because the gradients cannot be calculated in this random sampling process, model training through backpropagation becomes impossible; therefore, a reparameterization trick is applied (refer to Ref. [32] for more information about this trick). The latent variable $z$, which is assumed to take the form of the Gaussian distribution, can be expressed as follows:

$$z_i = \mu_i + \sigma_i \odot \epsilon \text{ and } \epsilon \sim N(0,1). \tag{5}$$

As presented earlier, the loss function for training the VAE model consists of a reconstruction error and the KL

divergence of the approximation function from the prior. The reconstruction error can be calculated using the binary cross entropy between the original data and reconstructed data from the VAE (reconstructed data refers to data after the original data goes through the encoder and decoder of the VAE). Then, to calculate the $KL(q_\phi(z|x)||p_\theta(z))$ term, $q_\phi(z|x_i) \sim N(\mu_i, \sigma_i^2)$ and $p_\theta(z) \sim N(0,1)$ are assumed. Finally, the KL divergence in the loss function for the datapoint $x_i$ can be expressed as follows, where $J$ is the dimensionality of the latent variable $z$:

$$\mathcal{L}(\theta, \phi; x_i) = \frac{1}{2}\sum_{j=1}^{J}(1 + \log((\sigma_i^j)^2) - (\mu_i^j)^2 - (\sigma_i^j)^2). \quad (6)$$

Backpropagation with respect to the loss function (composed of reconstruction error and KL divergence) enables the decoder of the VAE to be trained to generate realistic data. The flowchart of the VAE is shown in Fig. 1.

## 3. Inverse design optimization framework

### 3.1 Two-step deep learning approach

This section demonstrates the two-step deep learning approach, which connects the VAE and MLP models. First, the VAE was trained with the target performance distributions of the training data. When training has been completed, only the decoder part of the VAE model was used; it operates as a data generator that receives a low-dimensional latent variable and outputs a realistic high-dimensional target distribution. Then, the MLP was trained to predict the QoI and shape parameters from the target distribution. This regression process is more efficient and accurate compared with previous inverse design studies in that there is no iteration and theoretical/empirical assumptions (prediction was performed almost in real-time). In this study, these two deep learning models, the decoder of the VAE and MLP, were utilized sequentially; the decoder generates the distribution once it receives the latent variable, and the MLP outputs the QoI and shape parameters from this generated distribution. This structure refers to a two-step deep learning approach and its flowchart is shown in Fig. 2.

### 3.2 Target distribution optimization

The two-step deep learning approach allows mapping from the latent space of the target performance distribution to the QoI and shape parameters. Therefore, the optimization of the target distribution can be performed based on this approach. The inputs of the corresponding approach (latent variables) were used as the optimization variable, and the outputs (QoI and shape parameters) as the objective functions and constraints. In this study, because the shape parameters were incorporated into the target distribution optimization procedure, geometric constraints can be imposed directly as shape parameters. At the end of the optimization, numerical validation was performed at the optimum solutions by comparing the QoI predicted using the two-step approach and QoI calculated using the numerical solver. The inverse design optimization framework terminates when the differences in these values satisfy the error criterion. If not, the process described in *section 3.3* is repeated until it is satisfied.

### 3.3 Active learning and transfer learning

Since the VAE and MLP models are trained with initial training data, the optimization based on them is unlikely to meet the desired accuracy at once. Therefore, surrogate-based optimization studies usually add training data repeatedly to increase the accuracy of the surrogate model. This technique is called the active learning strategy and was adopted in this study for an accurate inverse design optimization framework [33, 34]. When the error criterion at optimum solutions is not satisfied (as presented at the end of *section 3.2*), these solutions were added to the previous dataset and the deep learning models were trained again. In the training procedure based on data splitting, which splits the full dataset into training and test data, the designs newly added at every iteration in the active learning process should be incorporated in the training data to maximize the efficiency of its process. This is because to reflect the newly added designs directly in the model training, they should be included in the training dataset, not the test dataset (test data is not used directly in model training). Then, optimization and numerical validation were performed based on these retrained models. This active learning strategy continues until the error between the QoI from the two-step deep learning approach and those from the numerical solver decreases so that they satisfy a predefined error criterion.

Although the active learning process was applied to effectively improve the accuracy of the framework, restarting model training with randomly initialized weights and biases at every iteration can severely degrade the computational efficiency. Moreover, in general, when adding new data via active learning, the model does not change significantly compared with that of the previous iteration, as only a small amount of data is added to the existing data. Therefore, this study used a parameter-based transfer learning strategy. This technique ensures that the weights and biases of the previously trained models are transferred to the models to be newly trained [35, 36]. Combining these two strategies, iterative active learning for model accuracy can be efficiently performed through transfer learning. The flowchart of the inverse design optimization framework, which summarizes the contents of *sections 3.1–3.3*, is shown in Fig. 3.

## 4 Framework validation: optimization of the airfoil for wind turbine blade

The proposed inverse design optimization framework can be applied to all inverse design problems in any engineering field. Specifically, in aerospace engineering, inverse design is actively being applied to wind turbine design [37-39]; therefore, to verify the accuracy, effectiveness, and robustness of this framework, airfoil optimization of a megawatt-class wind turbine was chosen as the application. For the airfoil design of a wind turbine blade, structural and aerodynamic performances are mainly considered. In particular, the airfoil at the blade tip is known to be critical for the aerodynamic performance of the blade. This study aims to optimize the airfoil at the tip of the blade, mostly taking into account its aerodynamic properties (structural performance is indirectly considered by the airfoil area). Single-objective and multi-objective optimizations were performed to demonstrate the versatility of the proposed framework in various optimization problems. The following sections describe the optimization problems (*section 4.1*), architectures of the two-step deep learning models used in the inverse design optimization (*section 4.2*), and the results and discussion of the single-objective and multi-objective optimizations (*sections 4.3* and *4.4*).

### 4.1 Optimization of the airfoil of a wind turbine blade

This section presents the optimization problems of the airfoil of a wind turbine blade tip region. *Section 4.1.1* presents the validation of the numerical flow solver used in this study, and *section 4.1.2* presents the problem definitions for the optimization.

### 4.1.1 Flow solver

Xfoil is a two-dimensional panel code capable of viscous/inviscid analysis, and it derives accurate results in a very short time when used appropriately [40]. Because a wind turbine operates at relatively low Reynolds numbers, numerous wind turbine airfoil design studies have used this solver [41-43]. In this study, Xfoil was adopted to calculate the aerodynamic QoI with reasonable computational cost. It should be noted that the proposed inverse design optimization framework can be coupled with any numerical solver with arbitrary QoI.

Although Xfoil is a well-known and widely used solver, we performed solver validation using experimental results [44]. The experimental data are based on the GA(W)-1 airfoil with Reynolds of $6.3 \times 10^6$, Mach of 0.15, and angle of attack of $8.02°$ (the flow conditions of the validation are intended to be similar to those of subsequent airfoil optimizations). It was confirmed that this flow solver is appropriate for the wind turbine airfoil optimization performed in this study as it predicts a pressure distribution almost identical to that from experiments (Fig. 4).

### 4.1.2 Optimization problem definitions

Before performing the optimization, the optimization problems should be defined first. There are numerous parameterization methods for representing the airfoil shape, such as PARSEC, B-spline, and class-shape transformation (CST) [45-47]. In this study, PARSEC parameters were adopted as they were originally introduced owing to their close relationship with the aerodynamic characteristics [48]. There are 11 PARSEC variables; six of them were used as shown in Fig. 5, namely $R_{L.E.}$ (leading-edge radius), $X_{up}$ (x-coordinate of the upper crest), $Z_{up}$ (z-coordinate of the upper crest), $X_{low}$ (x-coordinate of the lower crest), $Z_{low}$ (z-coordinate of the lower crest), and $Z_{T.E.}$ (z-coordinate of the trailing-edge). These variables were selected from the prior sensitivity test, which demonstrated that they have a significant impact on the flow characteristics, whereas the other five PARSEC variables have little impact. The corresponding multi-dimensional design space to be explored in the optimization process is summarized in Table 1 (the baseline airfoil shape is selected as the median value of each variable's range).

**Table 1.** Design space of the six airfoil shape parameters: the baseline airfoil is selected as the median value of each range

| Design variables | Lower bound | Baseline | Upper bound |
|---|---|---|---|
| $R_{L.E.}$ | 0.015 | 0.0275 | 0.04 |
| $X_{up}$ | 0.3 | 0.375 | 0.45 |
| $Z_{up}$ | 0.09 | 0.12 | 0.15 |
| $X_{low}$ | 0.3 | 0.375 | 0.45 |
| $Z_{low}$ | -0.15 | -0.12 | -0.09 |
| $Z_{T.E.}$ | -0.02 | 0 | 0.02 |

When the airfoil shape is determined using PARSEC parameters, flow analysis proceeds. Xfoil is executed under predefined settings (including the flight conditions), and the aerodynamic QoI that will be used as objective functions or constraints in the optimization are obtained. The corresponding flight conditions, objective functions, and constraints for single-objective and multi-objective airfoil optimizations are summarized in Table 2. In the single-objective optimization, the objective is to maximize the lift to drag ratio (L/D), which is the most crucial factor for the aerodynamic efficiency. Furthermore, some constraints are considered to exclude undesirable performance [49]: the drag should be less than the baseline value, the pitching moment coefficient at c/4 (c is the chord length) should be greater than the specific value to limit blade torsion, and the area should be at least 90% of the baseline area to prevent serious degradation of the structural performance. Note that the geometric constraint (airfoil area in this case) can be directly imposed in this framework as QoI in the target distribution optimization process (PARSEC parameters can also be set as geometric constraints, such as $Z_{up} < 0.15$, but these were realized by limiting the design space in this study). In the multi-objective optimization, the two objectives are to maximize the L/D ratio and airfoil area, considering that the area represents the structural performance. Other constraints are the same as those in the single-objective optimization.

**Table 2.** Flight conditions, objective functions, and constraints for single-objective and multi-objective optimizations

| Flight conditions | Reynolds number | $6*10^6$ |
| | Mach number | 0.25 |
| | Angle of attack | 7° |
| Single-objective optimization | Objective function | Maximize L/D |
| | Constraints | $C_d$ < Baseline $C_d$ |
| | | $C_m$ > -0.08 |
| | | Area > 0.9*Baseline Area |
| Multi-objective optimization | Objective functions | Maximize L/D |
| | | Maximize Area |
| | Constraints | $C_d$ < Baseline $C_d$ |
| | | $C_m$ > -0.08 |

Then, design of experiments was performed to train the MLP and VAE models; the sampled initial design samples were used as the initial training data. Latin hypercube sampling was selected owing to its uniformity in the design space [50]. A total of 500 initial designs were sampled and two deep learning models were trained based on them. Finally, the optimization proceeds using the latent variables of the VAE as the optimization variables, and the QoI and shape parameters as the objective functions and constraints, as shown in Fig. 3. For single-objective optimization, GA was adopted owing to its efficient global exploration in discontinuous and multimodal problems [51]. For multi-objective optimization, the non-dominated sorting genetic algorithm-Ⅱ (NSGA-Ⅱ) was adopted to obtain the diversified Pareto solutions of both objective functions [52]. Both optimization algorithms were implemented using Python package pymoo [53]. When the first optimization ends with two-step deep learning models trained using 500 DoE designs, active learning with transfer learning strategy was repeated iteratively. Because there is only one optimal solution in the single-objective optimization, the single optimal solution is selected as the design to be infilled. On the other hand, there are several optimal solutions for the multi-objective optimization (Pareto solutions). In this case, the leftmost, middle, rightmost designs in the Pareto solutions are selected to be infilled in order to increase the overall accuracy of the Pareto solutions. These criteria for infilling can be determined arbitrarily by the engineer (the number of points to be added for each iteration and their distribution in the Pareto frontier can be determined by the engineer as appropriate). Through these iterative procedures, the framework consisting of the two deep learning models satisfies the given error criterion. Again, note that the proposed framework can be applied to any inverse design problem (the design shape, corresponding shape parameters, QoI, numerical solver, and optimizer can be selected arbitrarily).

### 4.2 Architectures of the two-step deep learning models

In the two-step approach, the MLP model was trained to take the $C_p$ distribution as input and output QoI and shape parameters. First, all airfoils were discretized to share 199 identical x-coordinates: they were extracted using a two-sided hyperbolic tangent distribution function (where the spacing at the leading-edge and trailing-edge was constrained as 0.001c and 0.005c, respectively) [54] from NACA 0012 airfoil. Then, the pressure coefficients of the corresponding points were used as the input of the MLP. For the output, six shape parameters ($R_{L.E.}$, $X_{up}$, $Z_{up}$, $X_{low}$, $Z_{low}$, and $Z_{T.E.}$) and four QoI (L/D, $C_d$, $C_m$, and area) were concatenated to form the 10 output nodes. Finally, the MLP has 199 input nodes, 10 output nodes (all the inputs and outputs are normalized), and two hidden layers with 100 nodes: the MLP with the corresponding hidden layers was found to have sufficient accuracy for the regression in this problem. Then, LeakyReLU activation functions were applied to all the layers for nonlinearity. Adam was used as the optimizer with the mean square error loss function to train this MLP architecture, and the initial learning rate starts with 0.001. For the first iteration in active learning, 500 initial samples were split into training and test data in the ratio of 8:2 (the same ratio was also used to train the VAE). A total of 30000 epochs with a mini-batch size of 100 were performed using a scheduler that multiplies the learning rate by 0.8 for every 3000 epochs. Because there were no weight or bias values to be used as a reference in the first iteration, they were initialized using He initialization [55]. Then, active and transfer learning were performed. Interestingly, during the MLP training, it was observed that if the parameters of all layers from the previous model were passed, the training terminates without any meaningful change from the previous model. This is because the number of newly added designs is small compared with that of existing training data, and their effect on the loss function is negligible. Therefore, active learning, which increases the accuracy by appending previous optimum solutions to the current training data, becomes meaningless as their properties are not fully reflected in the current model. In this study, the parameters of the last layer of the MLP were initialized using He initialization, whereas those of other layers were transferred from the previous model (in other words, transfer learning was applied except for the last layer). For subsequent iterations, the required total epochs decreased to 10000 with the same initial learning rate and scheduler as the first iteration, owing to these transfer learning techniques. As a result, training the MLP in the first iteration using a personal computer (Intel Core i7-8700 3.2 GHz with 16 GB 2400 MHz DDR4 RAM) took 508 s, and subsequent iterations took an average of 186 s per iteration using Python package PyTorch [56]. The fact that subsequent iterations during active learning require a much shorter training time than the first iteration emphasizes the effect of applying the transfer learning technique in this study: by transfer learning technique, active learning can be efficiently performed.

For the VAE model, the 199 $C_p$ data previously inputted into the MLP were used as inputs and outputs (VAE has the same input and output). The 199-dimensional input data was reduced to four-dimensions using an encoder with hidden layers of 120, 60, and 30 nodes. Herein, the four-dimensions represent distribution parameters for random sampling in the latent space: two for the mean and the other two for the standard deviation of the Gaussian distribution. From these parameters, random sampling was performed and the dimension was reduced to a two-dimensional latent space. In this random sampling process, to sample a $J$-dimensional latent variable $z$, the same number of $\{\mu, \sigma\}$ sets is required as described in *section 2.2*, and $J$ equals 2 in this study. These two-dimensions were again reconstructed to 199-dimensions using a decoder with hidden layers of 30, 60, and 120 nodes. The corresponding architecture of the VAE is shown visually in Fig. 6. As in the MLP, the LeakyReLU activation function and Adam optimizer with mean square error loss function were used. The initial learning rate starts with 0.001, and a total of 30000 epochs were performed with a mini-batch size of 100 and a scheduler multiplying the learning rate by 0.5 for every 5000 epochs. In contrast to the MLP, the situation in which the learning process terminates without reflecting information of newly added designs is hardly observed in the VAE. Therefore, the parameters of all the layers from the previous iteration were transferred to the next iteration (in other words, transfer learning was applied to all layers). For subsequent iterations, a total of 10000 epochs were performed with the same initial learning rate and scheduler as the first iteration. As a result, training the VAE in the first iteration took 627 s, and subsequent iterations took an average of 226 s per iteration (again, the efficiency of transfer learning can be verified). All the hyperparameters of the MLP and VAE mentioned in this section were applied identically in the single-objective and multi-objective optimizations.

### *4.3 Single-objective optimization results and discussion*

Active learning of the single-objective optimization satisfies the error criterion (the error of the objective function at optimum solutions should be lower than 1%) after the 24 infilling iterations. The convergence history of optimization with active learning can be observed in Fig. 7. The objective function starts at approximately 65 and increases gradually, reaching a value of approximately 72 after 24 iterations. After that, no better optimal point was found. Subsequent analysis of the single-objective optimization results is based on the trained VAE and MLP at iteration 24 (total learning time for 24 iterations is (508+627) + (186+226) * 24 = 11,023 s).

The final optimal airfoil shape is shown in Fig. 8, and its QoI (objective function and constraints) values are summarized in Table 3. Its objective function (L/D) has a value of 72.22, which increased by 39% compared with the baseline. However, this value is just a prediction from the MLP model, and it is not certain whether the L/D calculated by Xfoil will have this value. Therefore, the QoI from Xfoil were compared with the predicted values from the MLP model. It was confirmed that the four QoI values have an error (between predicted and calculated) of less than 1%, and all the imposed constraints were satisfied. Note that the airfoil area satisfies the constraint imposed with little margin (0.6%), whereas other constraints (drag and pitching moment) were satisfied with some margin. Additionally, numerical validation was performed to verify whether the optimal airfoil has the $C_p$ distribution generated by the VAE model. Fig. 9 shows that the $C_p$ generated by the VAE and that calculated using Xfoil are almost indistinguishable. These results validate the accuracy of the MLP model in the two-step approach for single-objective optimization.

**Table 3.** Summary of the QoI of the optimum solution

|  | L/D | $C_d$ | $C_m$ | Area [m$^2$] |
|---|---|---|---|---|
| Baseline | 51.93 | 0.01436 | -0.0040 | 0.1574 |
| Optimum predicted | 72.22 | 0.01301 | -0.0173 | 0.1420 |
| Optimum calculated | 71.52 | 0.01305 | -0.0172 | 0.1426 |
| Error between predicted and calculated [%] | 0.98 | -0.28 | 0.72 | -0.46 |
| Comparison with baseline [%] | 39.08 | -9.40 | 332.50 | -9.82 |

Then, validation of the VAE model was performed. The VAE model reduces 199-dimensional $C_p$ distribution data to a two-dimensional latent space and reconstructs it to 199-dimensional data. The trained decoder in this study was used as a data generator that receives a two-dimensional latent variable and creates 199-dimensional $C_p$ distribution data. However, if the generated distribution cannot represent the overall training data or has completely different characteristics from them, the optimization results from this generator becomes inaccurate and inefficient. Therefore, the generated $C_p$ distributions by the trained VAE decoder were analyzed. Fig. 10 shows 50 randomly selected $C_p$ distributions from the 500 initial training $C_p$ data (black lines, Fig. 10a) and 50 randomly generated $C_p$ distributions by the decoder (red lines, Fig. 10b). Herein, the following points are confirmed. First, the data generated using the decoder covers the range of the training data well. Moreover, although no other technique was applied to smoothen the $C_p$ distribution during the VAE training, the decoder successfully generated smooth distributions indistinguishable from the training data. This supports the reason for adopting a VAE in this framework instead of a GAN; the VAE generates sufficiently realistic data (continuous $C_p$ in this case) without adopting auxiliary layers or filters to ensure the reality (continuity in this case) of the data. Second, from the generated $C_p$ distributions, we can also identify their dominant features. As the black dashed box indicates, significant shape differences are observed near the suction peak (near the leading-edge of the airfoil's upper surface), whereas most differences in the other regions are just slight shifts in the $C_p$ values. From the fact that these dominant features near the suction peak are also observed in the training data (Fig. 10a), it can be concluded that the VAE model successfully learned the characteristics of the training data. In summary, since we have confirmed that the data generated by the VAE can be well representative of the training data, the trained VAE model will perform successfully as a data generator in this framework.

### 4.4 Multi-objective optimization results and discussion

In *section 4.3*, it was confirmed that the proposed inverse design optimization framework yields outstanding results when single-objective optimization is performed. In this section, results of multi-objective optimization are presented to ensure the universality of the framework in various optimization problems. In the previous single-objective problem, which uses L/D as an objective function and airfoil area as a constraint, the area of the optimum solution barely satisfied the imposed constraint with little margin (0.6%). This indicates that a potential increase in the objective function L/D is suppressed by the area constraint. Therefore, in the multi-objective optimization, these two QoI were set as objective functions to consider their trade-off relationships (aerodynamic and structural performance). Refer to Table 2 for the problem definition of multi-objective optimization.

The active learning process of the multi-objective optimization converged after 59 iterations. Accordingly, 677 data consisting of initial 500 points and 177 infilled points were used (the total learning time for 59 iterations is (508+627) + (186+226) * 59 = 25,443 s). The resultant Pareto frontier of the two objective functions is shown in Fig. 11. We observed a discontinuity in the middle of the calculated Pareto solutions, and it is concluded that this is due to $C_d$ constraint violation (the Pareto frontier from the optimization without the $C_d$ constraint is smoothly connected: this result is shown in "Appendix"). Among the Pareto solutions, six designs (A1-A3 and B1-B3) were selected for further analysis, and their airfoil shapes are shown in Fig. 12. Herein, it can be seen that along with the Pareto frontier, airfoil shapes of selected six designs change sequentially: as the performance criterion changes from area to L/D (from A1 to B3), the airfoil thickness gradually decreases. Xfoil was performed on these six designs to estimate the error between the QoI from framework prediction and solver calculation, as in single-objective optimization. The corresponding results are summarized in Table 4: the errors between the QoI predicted using the framework and those obtained using Xfoil are within a reasonable range. In particular, A2 and B1 have quite large errors because there are few points added nearby in the active learning process (the percentage errors of $C_m$ are also large, but this is due to their scale). The accuracy near these points can be increased by adding points close to them (it is up to the engineer where to infill points in the Pareto solutions). The accuracy of the trained MLP model was evaluated by verifying that the six selected designs actually have $C_p$ distributions generated by the VAE. Fig. 13 shows that the MLP is accurate enough in that the $C_p$ calculated using Xfoil and that generated using the VAE are almost indistinguishable, as in single-objective optimization.

**Table 4.** Summary of the QoI of six selected Pareto solutions

|  |  | L/D | Area [m$^2$] | $C_d$ | $C_m$ |
|---|---|---|---|---|---|
|  | Baseline | 51.93 | 0.1574 | 0.01436 | -0.0040 |
| A1 | Predicted | 34.56 | 0.1844 | 0.01436 | 0.0513 |
|  | Calculated | 34.26 | 0.1840 | 0.01445 | 0.0514 |
|  | Error [%] | -0.85 | -0.21 | 0.63 | 0.17 |
| A2 | Predicted | 42.16 | 0.1781 | 0.01434 | 0.0318 |
|  | Calculated | 40.63 | 0.1789 | 0.01461 | 0.0343 |
|  | Error [%] | -3.64 | 0.44 | 1.91 | 7.71 |
| A3 | Predicted | 49.86 | 0.1697 | 0.01423 | 0.0124 |
|  | Calculated | 49.52 | 0.1697 | 0.01428 | 0.0124 |
|  | Error [%] | -0.68 | 0.03 | 0.35 | 0.19 |
| B1 | Predicted | 54.13 | 0.1585 | 0.01436 | -0.0027 |
|  | Calculated | 55.22 | 0.1579 | 0.01442 | -0.0075 |
|  | Error [%] | 2.01 | -0.38 | 0.43 | 179.92 |
| B2 | Predicted | 64.42 | 0.1493 | 0.01433 | -0.0256 |
|  | Calculated | 64.99 | 0.1488 | 0.01445 | -0.0333 |
|  | Error [%] | 0.89 | -0.36 | 0.86 | 30.05 |
| B3 | Predicted | 75.99 | 0.1396 | 0.01307 | -0.0387 |
|  | Calculated | 76.32 | 0.1396 | 0.01299 | -0.0383 |
|  | Error [%] | 0.43 | -0.05 | -0.63 | -1.12 |

In this framework, the latent variable goes through the VAE decoder and MLP sequentially to predict the QoI values. Based on this two-step approach, an optimization technique is applied to obtain a latent variable that maximizes/minimizes the QoI. Therefore, for efficient optimization through this two-step deep learning approach, two-step deep learning models should learn the precise correlation between the two spaces (the latent space and QoI space). However, in real engineering problems, this can be difficult due to abrupt changes of physical conditions such as shock waves. When the mapping is inaccurate, the efficiency of the optimization technique based on their mapping will be significantly undermined. In this context, this study investigated the mapping between the latent space and QoI through heatmaps to verify how they are correlated, as shown in Fig. 14: heatmap of L/D and area is shown in Fig. 14a and Fig. 14b, respectively. In this figure, it can be observed that the latent space is mapped continuously to both objective functions, indicating how optimization in this framework could be performed successfully. On the other hand, sharp changes in both objectives are detected when $z_2$ has a value of approximately -0.55 (as indicated by the yellow squares). To investigate the changes occurring in the flowfield across this boundary, a total of 12 points were extracted nearby, as shown at the top of Fig. 14. The nomenclatures of these 12 points are summarized in Table 5.

**Table 5.** Nomenclatures of twelve points extracted to investigate the sharp changes in the QoI heatmaps

|  |  | $z_1$ | | |
|---|---|---|---|---|
|  |  | -2 | -1.5 | -1 |
|  | -0.4 | L4 | M4 | R4 |
|  | -0.5 | L3 | M3 | R3 |
| $z_2$ | -0.55 | *Boundary of rapid change in the QoI heatmap* | | |
|  | -0.6 | L2 | M2 | R2 |
|  | -0.7 | L1 | M1 | R1 |

To find the reason for the sharp change in the QoI at $z_2 \approx$ -0.55, the $C_p$ distributions generated through VAE from 12 selected points' latent variables are shown in Fig. 15 (in this figure, the horizontal and vertical axes of the plots represent x/c and $C_p$, respectively, and they are scaled to the same range for the comparison). In this figure, the local leading-edge suction peaks on the lower curve are not observed in the $C_p$ plots of $z_2$ > -0.55, whereas they are observed in the plots of $z_2$ < -0.55. Therefore, it can be inferred that $z_2 \approx$ -0.55 is the boundary between the presence and absence of the leading-edge suction peak. In this regard, it is known that the sharper the leading-edge of the airfoil, the larger the suction peak behind the leading-edge (as the leading-edge radius decreases, the angle at which the flow should bend increases, thereby increasing the suction in order to attach the flow to the airfoil surface [57]). Therefore, the trends in the leading-edge radius ($R_{L.E.}$) are depicted in Fig. 16 to verify whether the rapid changes in $C_p$ are accompanied by the changes in $R_{L.E.}$. Indeed, as $z_2$ decreases (as the number corresponding to the second character of the nomenclature decreases), the leading-edge radius decreases, which is consistent with the prior knowledge. Moreover, in all cases of L, M, and R, there are noticeable gaps between 2 and 3. Considering that nomenclatures 1, 2, 3, and 4 are equally spaced in the $z_2$-direction, it can be concluded that a rapid change in the leading-edge radius at the boundary between 2 and 3 ($z_2 \approx$ -0.55) leads to a sudden change in the trend of the leading-edge suction. Additionally, the six points selected from the Pareto solutions in Fig. 11 were scrutinized in a similar way. In Fig. 13, the leading-edge suction peaks are not found in the $C_p$ distributions of A1-A3, but are found in B1-B3.

In fact, when the latent variables of these designs are visualized as shown in Fig. 14b, the two groups (A1-A3 and B1-B3) are separated by a boundary of $z_2 \approx$ -0.55. In summary, by analyzing the heatmaps, mapping between the latent space and QoI using the two-step approach is verified to be generally continuous. Additionally, this mapping is verified to accurately reflect the rapid changes in the QoI, which occur frequently in real-world engineering applications. This flexibility of the two-step deep learning approach enables the optimization to be performed efficiently owing to continuous and accurate mapping between the optimization inputs and outputs, highlighting the capability of the proposed inverse design optimization framework.

## 5 Conclusions

This study proposed a novel inverse design optimization framework with a two-step deep learning approach, which refers to consecutive coupling of VAE and MLP. Herein, the VAE generates a realistic target distribution and MLP predicts the QoI and shape parameters from the generated distribution. Then, the target distribution was optimized based on this two-step approach. To increase the accuracy, we used active learning to retrain the models with newly added designs. Herein, transfer learning was coupled to reduce the computational cost required for retraining. These techniques increase the accuracy of the framework with efficient computational resources. Finally, the limitations of previous inverse design studies can be eliminated through the proposed framework as follows:

1) The conventional inverse design process is substituted with the MLP surrogate model so that the iterations coupled with the flow solver are not required.

2) From the target distribution, the MLP surrogate model not only predicts the design shape, but also the QoI. Therefore, theoretical/empirical assumptions are not required for predicting the QoI, and the geometric constraints can be handled directly in the target distribution optimization process.

3) For the inverse design optimization, realistic target performance distributions are generated using a VAE deep generative model so that the loss of the diverse representation capacity due to the parameterization of the distribution is mitigated, and excessive constraints for the realistic distribution are not required.

The proposed framework was validated using two constrained optimization problems: single-objective and multi-objective airfoil optimizations of the tip region of a megawatt-class wind turbine blade. In the single-objective optimization, the prediction accuracy of the trained MLP model and the validity of the trained VAE model for generating realistic data were verified. In the multi-objective results, continuous mapping between the inputs and outputs of the framework was verified, which enables successful optimization through the two-step approach. Furthermore, this mapping was confirmed to accurately reflect the rapid changes in the QoI, which occur frequently in real-world engineering applications. In summary, the results of the optimizations show that the proposed inverse design optimization framework via a two-step deep learning approach is accurate, efficient, and flexible enough to be applied to any other inverse design problem.

Considering that this novel framework can be coupled with any numerical solver with arbitrary design shape and QoI, it can be easily applied and extended to various engineering fields. Moreover, the deep learning models in the two-step approach can be replaced by other suitable alternatives: the VAE by any data generator model and the MLP by any other regression model. For instance, the MLP, a widely used but quite simple model, was used in this study as the problem we investigated is not much complex to apply other advanced deep learning models. However, when the problem is complicated, other models such as convolutional neural networks or recurrent neural networks can be used instead of simple MLP model. This flexibility contributes to the versatility of the framework by allowing it to be utilized with any model suitable for the engineering problem to which it applies.

**Appendix**

The Pareto frontier from the optimization without the $C_d$ constraint is shown in Fig. 17: from the continuous Pareto solutions, it can be inferred that the discontinuity in the Pareto solutions in Fig. 11 was due to $C_d$ constraint violation. Also, as the constraint is eliminated, Pareto solutions without $C_d$ constraint have better performance near B1 design than Pareto solutions with $C_d$ constraint.

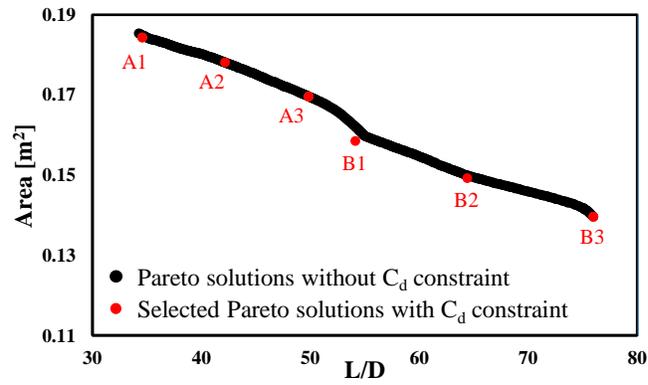

**Fig. 17** Pareto solutions of multi-objective optimization without $C_d$ constraint. For comparison with Fig. 11, six designs previously selected from the Pareto solutions with $C_d$ constraint are also shown

**Statements and declarations**

*Competing interests*

The authors have no conflicts of interest to declare.

*Authors' contributions*

Data curation [Sunwoong Yang]; Software [Sunwoong Yang]; Methodology [Sunwoong Yang]; Formal analysis [Sunwoong Yang]; Investigation [Sunwoong Yang]; Validation [Sunwoong Yang]; Visualization [Sunwoong Yang]; Writing – original draft [Sunwoong Yang]; Conceptualization [Sanga Lee]; Methodology [Sanga Lee]; Supervision [Sanga Lee]; Writing – review & editing [Sanga Lee]; Funding acquisition [Kwanjung Yee]; Project administration [Kwanjung Yee]; Resources [Kwanjung Yee]; Supervision [Kwanjung Yee]; Writing – review & editing [Kwanjung Yee]

**Figure captions**

**Fig. 1** Flowchart of the variational autoencoder (VAE)

**Fig. 2** Flowchart of the two-step deep learning approach

**Fig. 3** Flowchart of the inverse design optimization framework

**Fig. 4** Comparison of pressure distribution from Xfoil and experimental results in Ref. [44] (used airfoil is also visualized)

**Fig. 5** Shape parameters for representing the airfoil: six PARSEC parameters

**Fig. 6** Architecture of the VAE

**Fig. 7** Convergence history of single-objective optimization with active learning

**Fig. 8** Comparison of the baseline and optimum airfoil shape of single-objective optimization

**Fig. 9** Comparison of the generated and calculated pressure distribution of the optimum airfoil (baseline pressure distribution is also included)

**Fig. 10** Comparison of 50 randomly selected $C_p$ training data (black lines, **a**) and 50 generated $C_p$ distributions by the VAE (red lines, **b**). The black dashed box near the leading-edge of the lower curve indicates clear distinctions between generated distributions

**Fig. 11** Pareto solutions of multi-objective optimization. The discontinuity in the Pareto solutions is due to $C_d$ constraint violation

**Fig. 12** Airfoil shape comparison of six selected Pareto solutions

**Fig. 13** Comparison of generated and calculated $C_p$ distributions of six selected Pareto solutions

**Fig. 14** Heatmaps of two objective functions within the latent space: **a** L/D and **b** area. Twelve points were selected to investigate the rapid change at $z_2 \approx$ -0.55 (top), and the latent space of six selected Pareto solutions is shown in the heatmap of the area (**b**)

**Fig. 15** $C_p$ distributions of 12 selected points in Fig. 14

**Fig. 16** Trends in the leading-edge radius ($R_{L.E.}$) of 12 selected points in Fig. 14

**Figures**

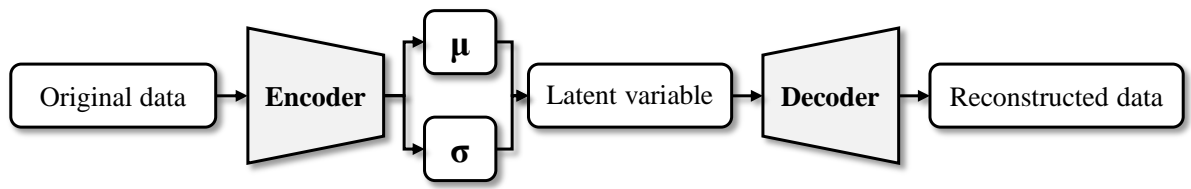

**Fig. 1**

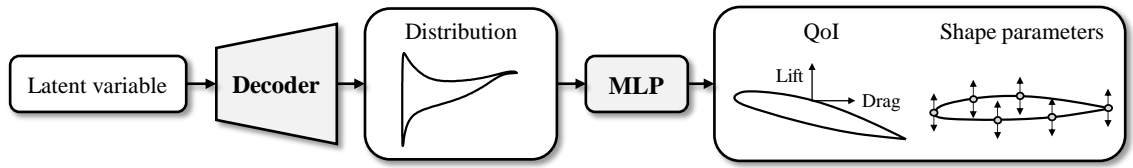

**Fig. 2**

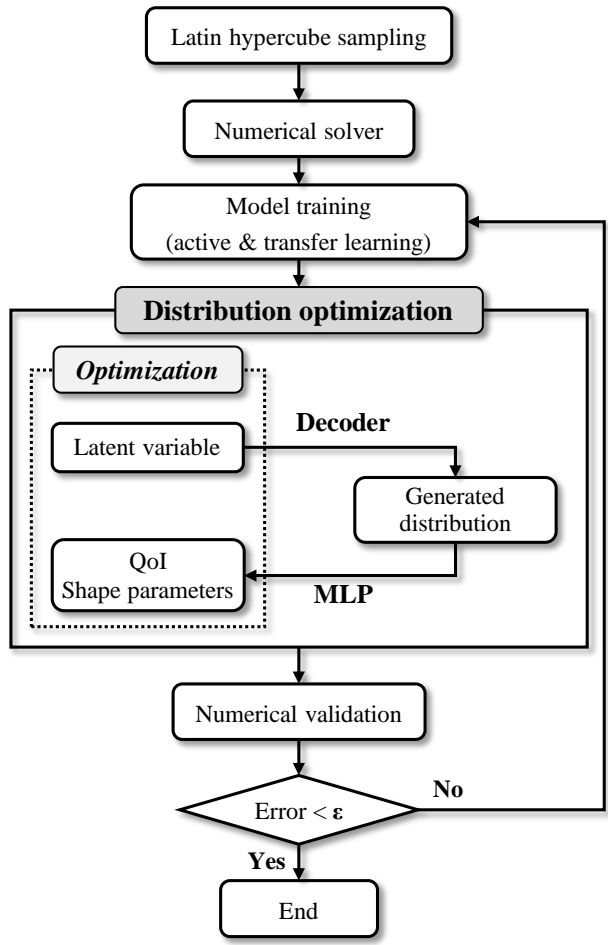

**Fig. 3**

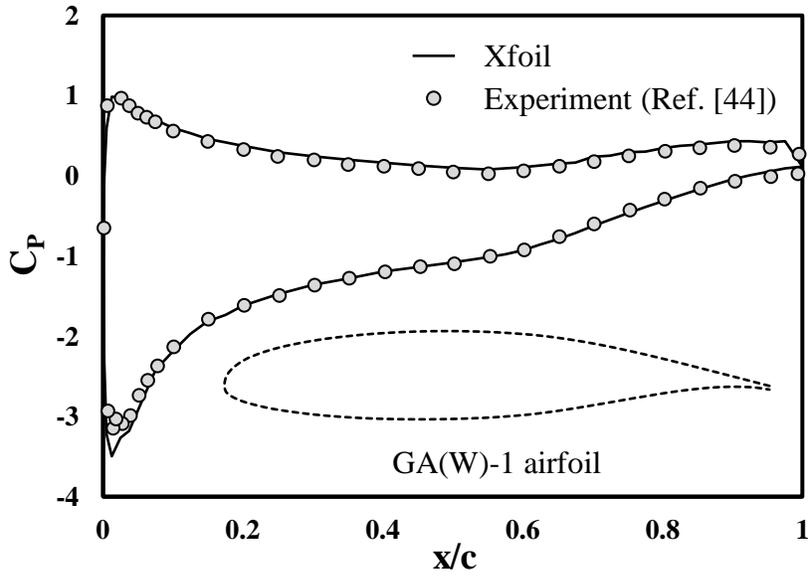

Fig. 4

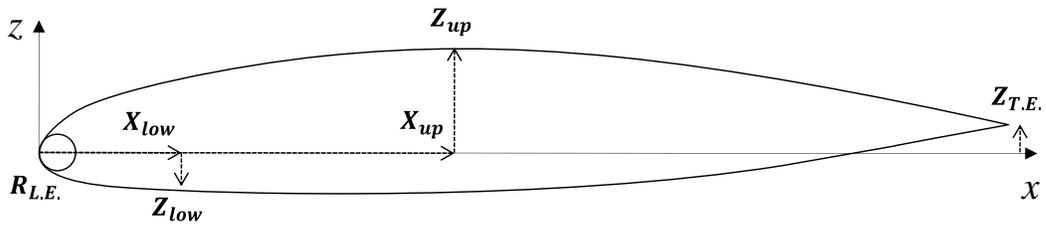

**Fig. 5**

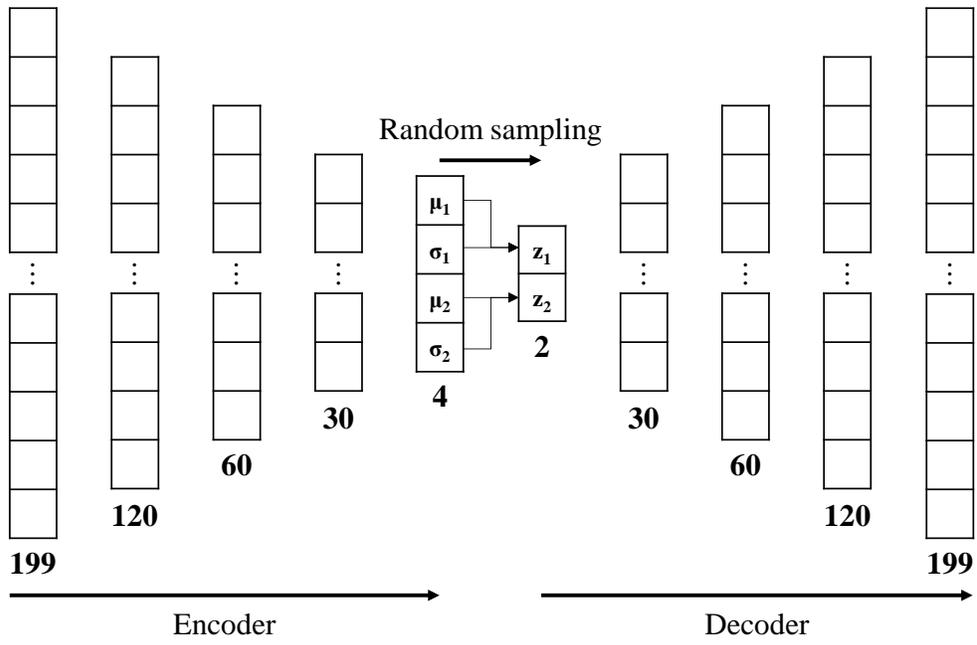

Fig. 6

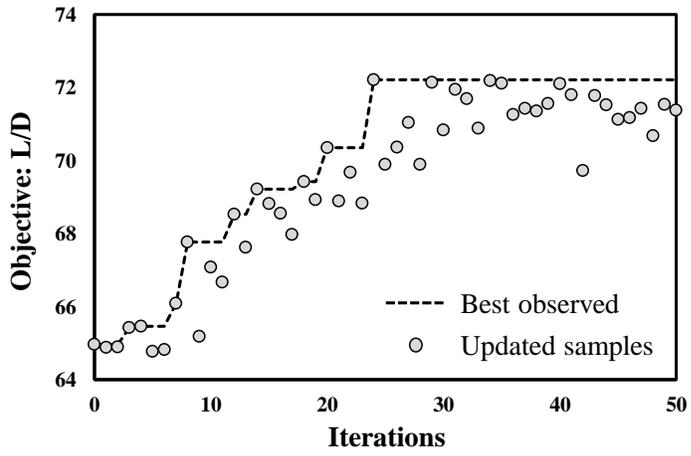

**Fig. 7**

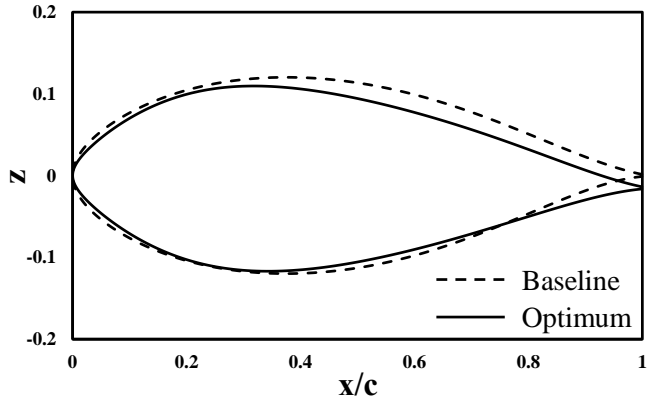

**Fig. 8**

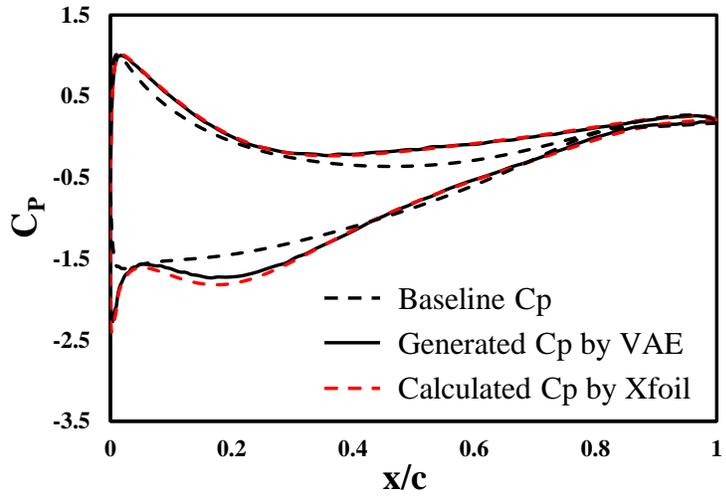

Fig. 9

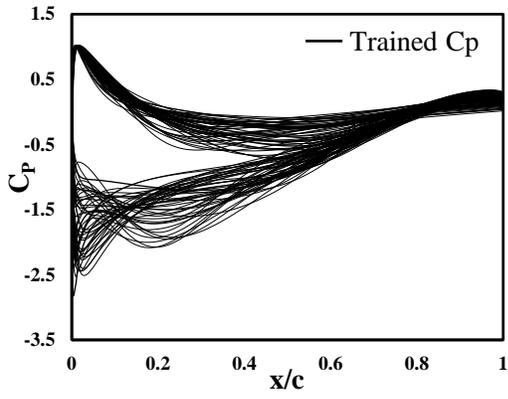 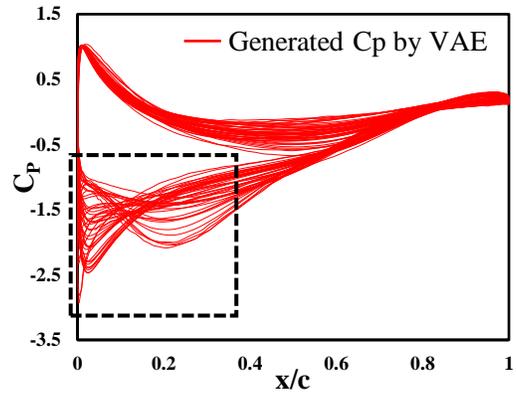

**(a)**                                    **(b)**

**Fig. 10**

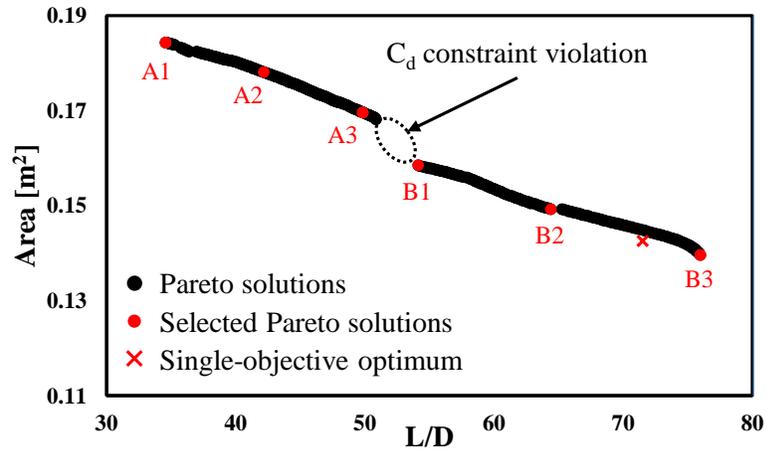

Fig. 11

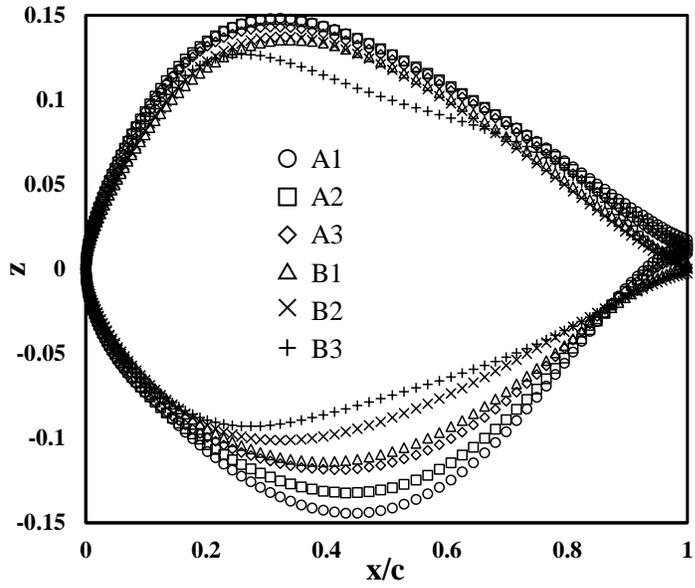

**Fig. 12**

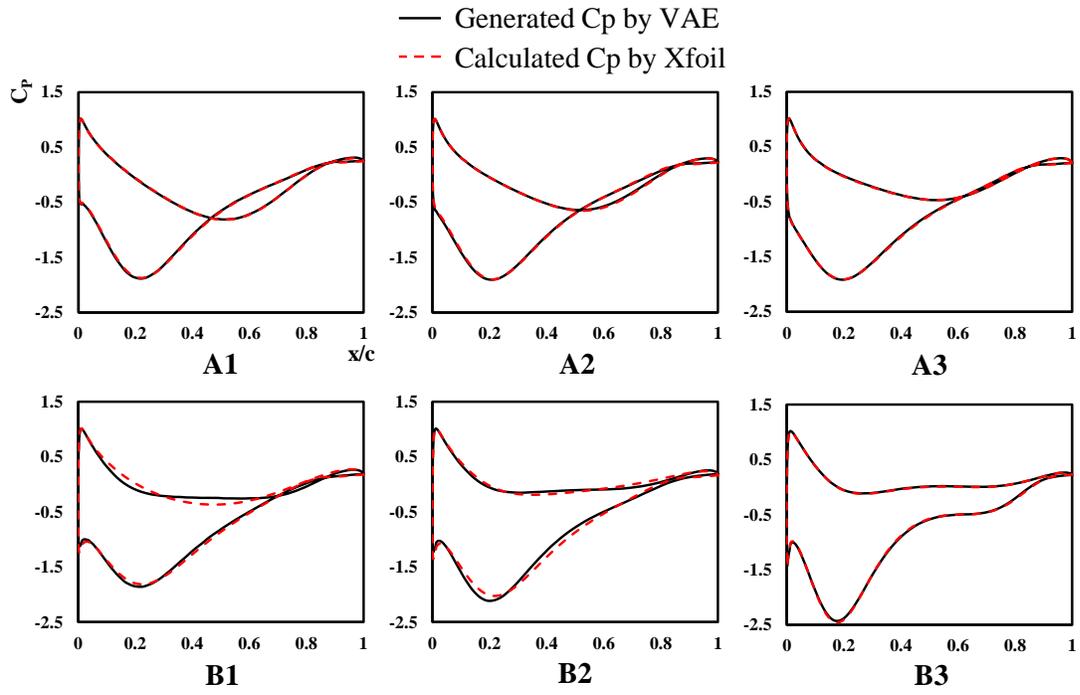

Fig. 13

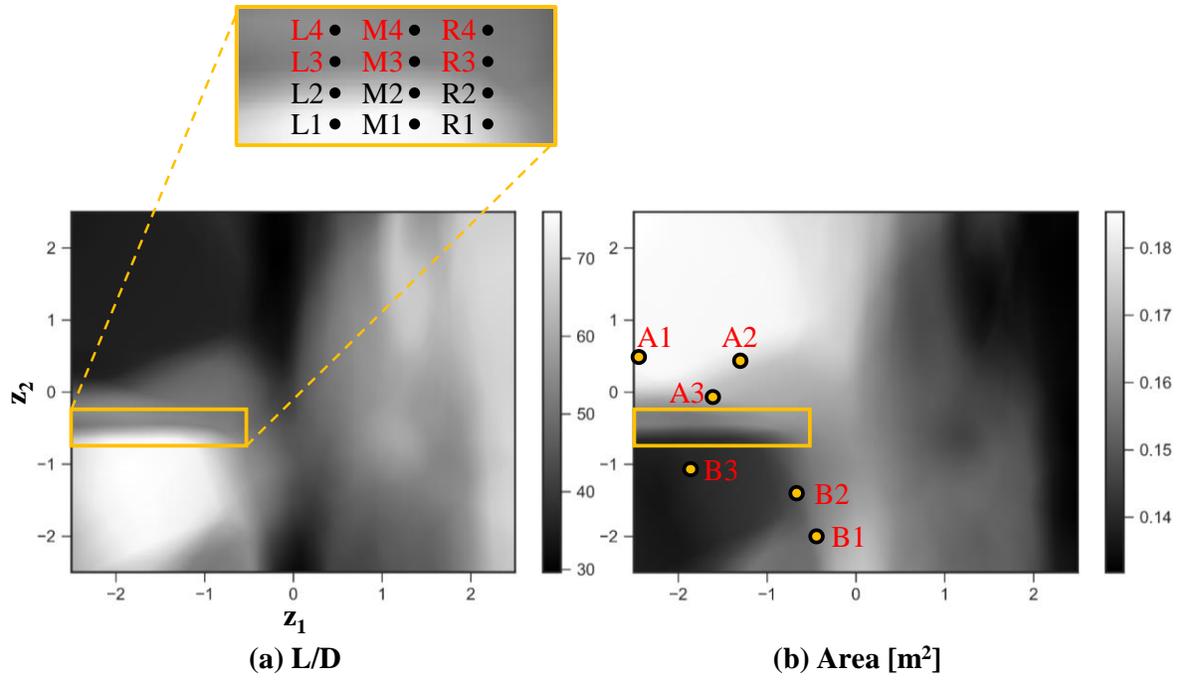

**(a)** L/D

**(b)** Area [m²]

Fig. 14

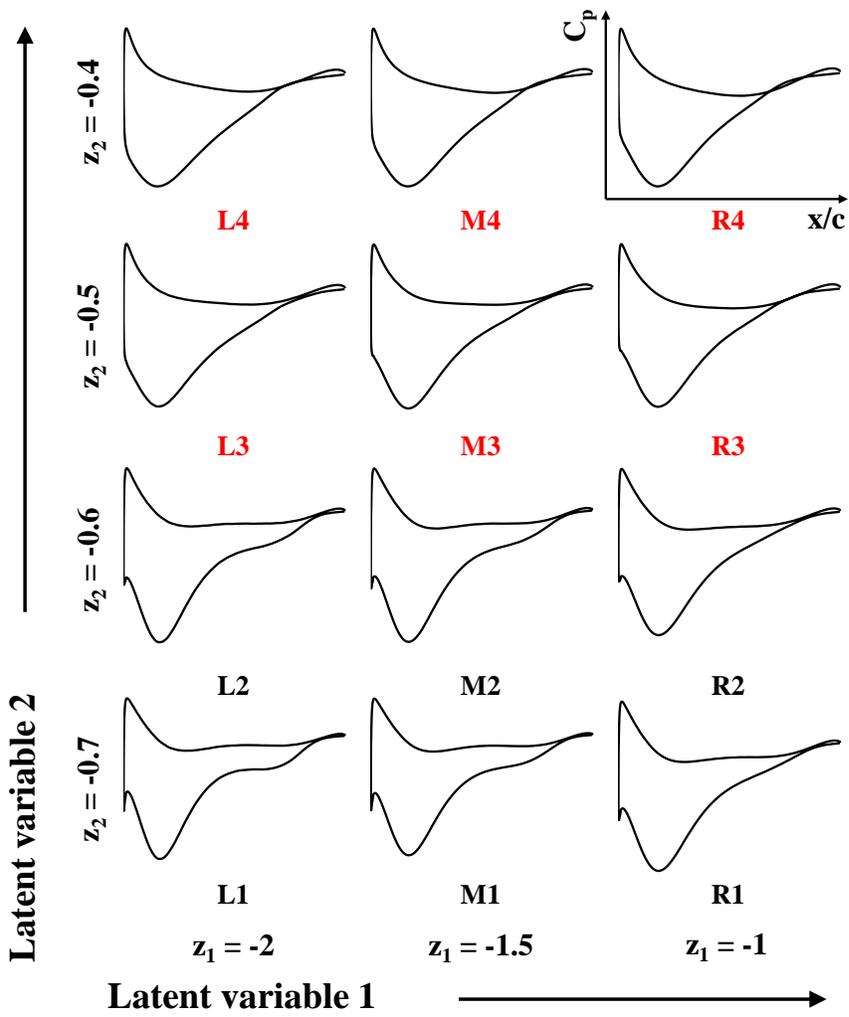

Fig. 15

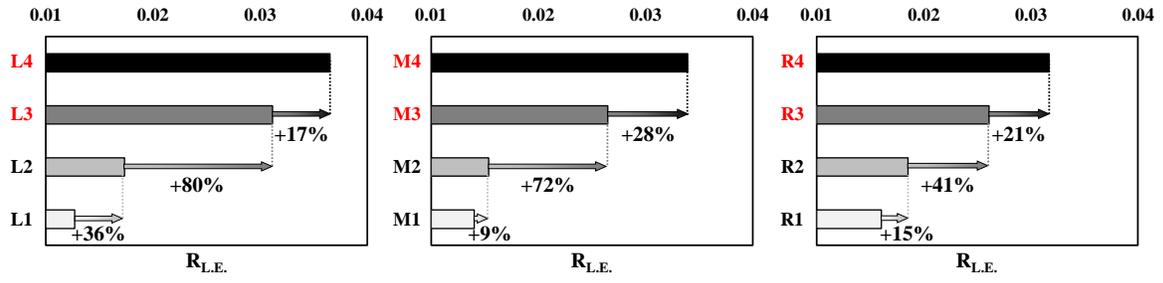

**Fig. 16**